\documentclass[11pt]{article}
\pdfoutput=1

\usepackage{graphicx}
\usepackage{tikz}
\usetikzlibrary{shapes}
\usepackage{caption}
\usepackage{multirow}
\usepackage{url}
\usepackage{algorithm2e}
\usepackage[numbers]{natbib}
\usepackage{microtype}
\usepackage{authblk}
\usepackage{booktabs}
\usepackage[colorlinks,bookmarksopen,bookmarksnumbered,citecolor=red,urlcolor=red]{hyperref}

\begin{document}

\title{Guetzli: Perceptually Guided JPEG Encoder}
\author{J.~Alakuijala}
\author{\href{mailto:robryk@google.com}{R.~Obryk}\footnote{robryk@google.com}}
\author{O.~Stoliarchuk}
\author{Z.~Szabadka}
\author{L.~Vandevenne}
\author{J.~Wassenberg}
\affil{Google Research Europe}
\maketitle

\pdfinfo{
	/Author (J.~Alakuijala, R.~Obryk, O.~Stoliarchuk, Z.~Szabadka, L.~Vandevenne, J.~Wassenberg)
	/Title (Guetzli: Perceptually Guided JPEG Encoder)
	/CreationDate (D:20170309000000)
	/Subject ()
	/Keywords ()
	/Creator ()
	/Producer ()
}

\begin{abstract} Guetzli is a new JPEG encoder that aims to produce visually
indistinguishable images at a lower bit-rate than other common JPEG encoders. It
optimizes both the JPEG global quantization tables and the DCT coefficient
values in each JPEG block using a closed-loop optimizer. Guetzli uses
Butteraugli~\cite{butteraugli}, our perceptual distance metric, as the source of
feedback in its optimization process. We reach a $29$-$45\%$ reduction in data size
for a given perceptual distance, according to Butteraugli, in comparison to other
compressors we tried. Guetzli's computation is currently extremely slow, which
limits its applicability to compressing static content and serving as a proof-
of-concept that we can achieve significant reductions in size by combining
advanced psychovisual models with lossy compression techniques. \end{abstract}

\section{Introduction}

Two thirds of the average web page size are spent on representations of images:
JPEGs, GIFs and PNGs; almost half of the image requests are JPEGs, which tend to
be much larger in byte size than PNGs and GIFs~\cite{web-stats}. Given that many
clients and particularly mobile clients are limited by transfer bandwidth, we
can make websites load faster by reducing the size of JPEG images. Standard JPEG
encoders allow trading off visual quality against size by tuning the
\emph{quality} parameter. In this work we look into how to reduce the size of
JPEG images without impacting the perceived visual quality of the images.

\begin{figure}
\includegraphics[width=\textwidth]{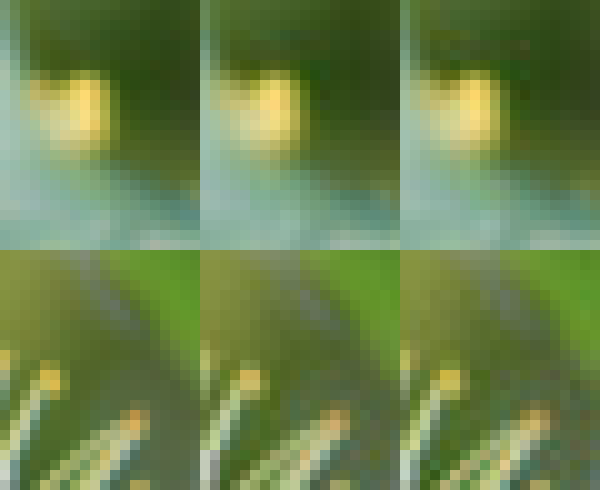}
\caption{Visualization of two details in an image, original in the left column,
Guetzli in the middle, libjpeg on the right. Libjpeg shows more ringing
artifacts than Guetzli.}
\label{fig:sample-artifact}
\end{figure}

We visually observed that JPEGs encoded with existing encoders typically have
inhomogeneous quality; they often exhibit disturbing artifacts only in a few
places on the image. Often areas close to sharp edges or lines exhibit more
visible artifacts (e.g. as in Fig.~\ref{fig:sample-artifact}). This led us to
think that further optimization is possible.  We assume that when an encoder
throws away information in an efficient manner, the JPEG image should start to
degrade roughly evenly everywhere when the degradation starts to become visible.
With Guetzli we attempt to cause a degradation in visual quality that is both
more homogeneous and yields smaller JPEG images.

\begin{figure}
\includegraphics[width=\textwidth]{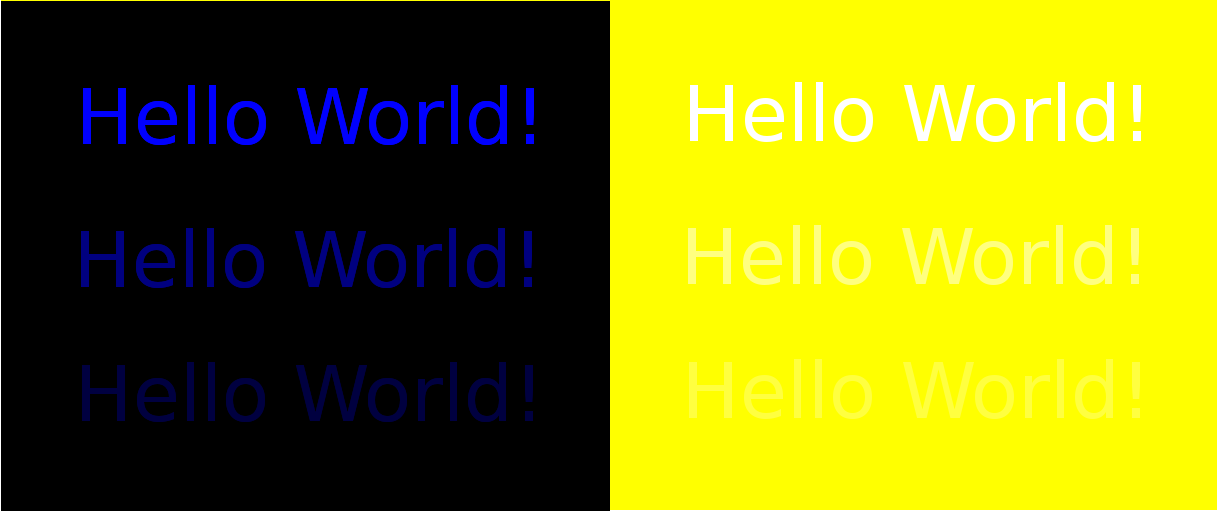}
\caption{Experiment with additive blue channel signal on black and yellow
backgrounds shows that blue changes are more difficult to see on the yellow
background. Receptors (cones) at retina receive the colors in such a way that
different components can mask changes in other components. Here, we show how
changes in the low intensity blue component are masked by the high intensity
levels in the red and green components. The same differences in blue are more
difficult to see against a yellow background than against the black background.
By using Butteraugli, Guetzli detects the lesser importance of blue on a yellow
background and stores it with less accuracy.}
\label{fig:blue-on-background}
\end{figure}

\begin{figure}
\includegraphics[width=\textwidth]{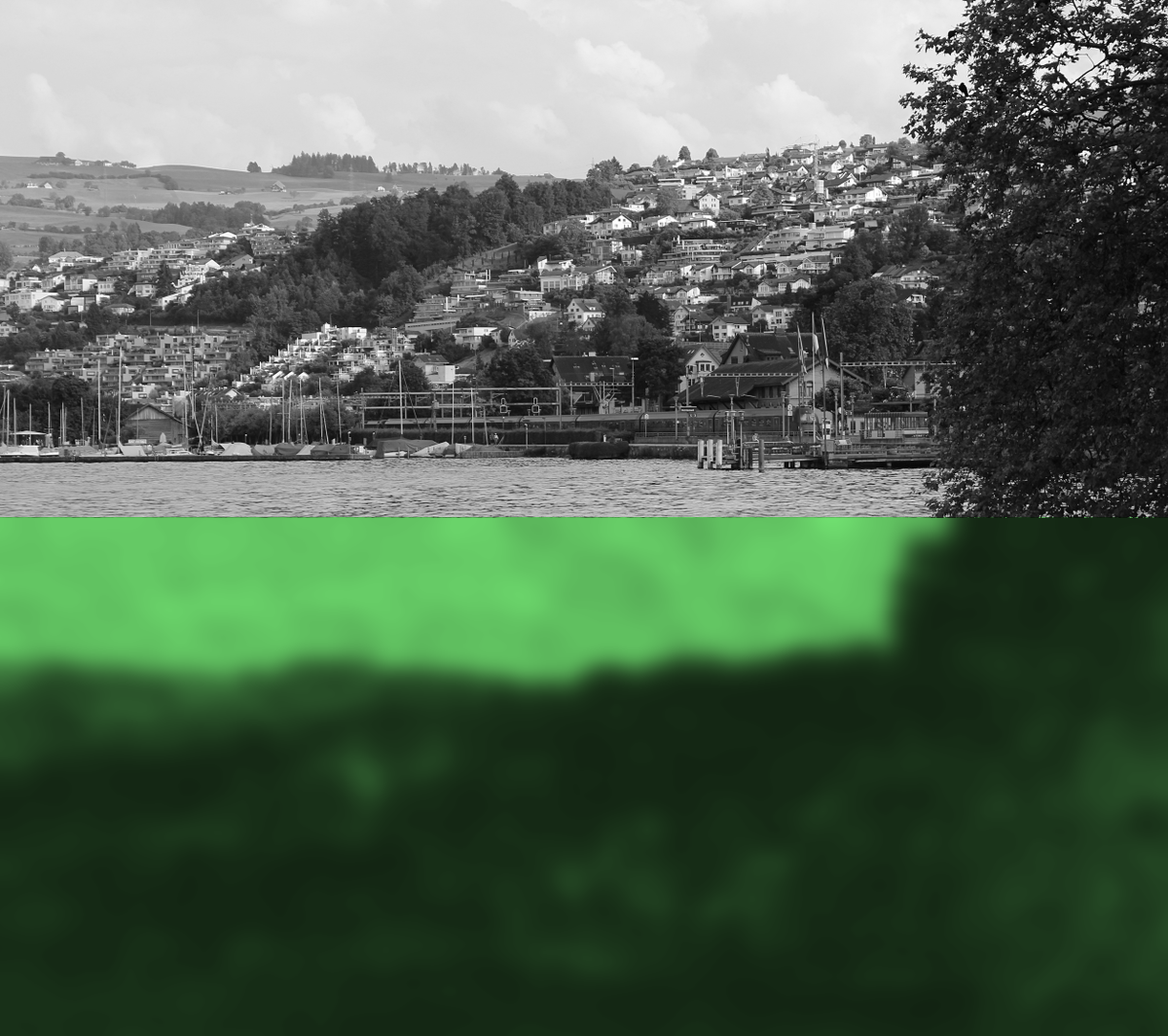}
\caption{Landscape photo on the top with its visual mask shown on the bottom.
Darker areas on the visual mask require less precise reproduction of details.
Visual masking allows areas of the photograph to be stored at different
accuracy, up to $6\times$ quantization difference for this image. According to this
visual masking model, the sky needs to be compressed with less loss than the
tree, lake and the buildings for a uniform experience of compression quality.
Guetzli computes two separate masking models -- one for low spatial frequency
color modeling and one for high spatial frequency color modeling. Both models
contain one mask for each dimension of the color space. The mask above is the
high spatial frequency intensity mask.}
\label{fig:visual-masking}
\end{figure}

Guetzli is an open source JPEG encoder~\cite{guetzli-source} that targets very
high perceptual qualities. It performs a closed-loop optimization, with feedback
provided by Butteraugli, our model of human vision~\cite{butteraugli}. Its goal
is to find the smallest JPEG which cannot be distinguished from the original
image by the human eye according to Butteraugli. Butteraugli takes into account
three properties of vision that most JPEG encoders do not make use of. First,
due to the overlap of sensitivity spectra of the cones, gamma correction should
not be applied to every RGB channel separately. There is some relationship
between e.g. amount of yellow light seen and sensitivity to blue light. Thus,
changes in blue in the vicinity of yellow can be encoded less precisely
(Fig.~\ref{fig:blue-on-background}). YUV color spaces are defined as linear
transformations of gamma-compressed RGB and thus are not powerful enough to
model such phenomena. Second, the human eye has lower spatial resolution in blue
than in red and green, and has next to no blue receptors in the high-resolution
area of the retina. Thus, high frequency changes in blue can be encoded less
precisely. Third, the visibility of fine structure in the image depends on the
amount of visual activity in the vicinity. Thus, we can encode areas with large
amount of visual noise less precisely (see example in Fig.~\ref{fig:visual-masking}).
In Guetzli we model all these aspects in a way that leads to
homogeneous loss in the image. We achieve this by guiding the encoder with
Butteraugli, our psychovisual metric.

In this document we describe the optimization approaches we use and
ones we have rejected, the iterative framework in which we apply those approaches,
show the results in comparison to other JPEG encoders, and finally discuss their
significance and further opportunities in image compression.

\section{Methods}

\begin{figure}
\includegraphics[width=\textwidth]{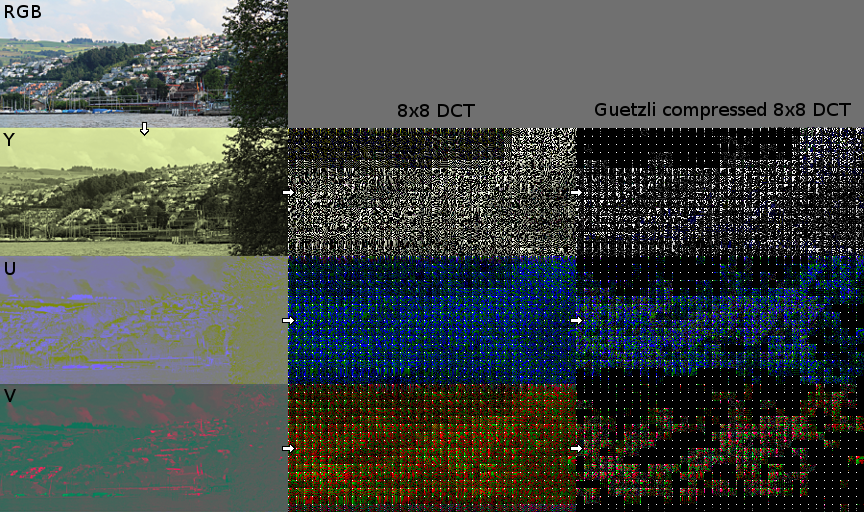}
\caption{ On the left hand, the landscape image on the top left hand corner is
decomposed into three YUV planes in a JPEG. Every $8\times{}8$ square is transformed into
DCT space, and the DCT values are quantized. In addition to quantizing, Guetzli
zeroes out small values aggressively.}
\label{fig:winnowing}
\end{figure}

JPEG encoding consists of converting an image to YUV colorspace, breaking it up
into blocks, transforming each block into frequency domain using DCT, quantizing
the resulting coefficients and compressing them losslessly
(Fig.~\ref{fig:winnowing}). Guetzli looks for possibilities to reduce the size
of the compressed representation without degrading the perceived visual quality.
This section describes the methods used to achieve that.

\subsection{Optimization opportunities}

Guetzli produces a compliant JPEG file, so the optimizations that can be
performed are limited strictly to the options available in this data format, and
even further limited to those that practical implementations support. We use
three options provided by the format: we tune the global quantization tables,
replace some DCT coefficients with zeroes and decide on using a mode in which chroma channels
are downsampled (YUV420). We have decided not to use other options, either
because we found them not to be beneficial, or because they cause other
undesirable effects.

The first optimization opportunity we make use of is changing the (global)
quantization tables to make the quantization coarser, which decreases the size
of the image (by decreasing the magnitude of stored coefficients). This is
similar to adjusting the quality parameter in a traditional JPEG encoder and
causes distortions in the whole image.

The second opportunity involves direct modification of the coefficients. We
replace some of the DCT coefficient values in each block with zeros. This
modification distorts the visual appearance of the block in question. Zeros are
RLE-encoded, so encoding a zero that occurs next to another zero costs virtually
nothing. Thus, replacing a coefficient with a zero, when there is a neighbouring
zero, reduces the encoded size by the size of that coefficient. Even if there is
no neighbouring zero, encoding of a zero is virtually always shorter than of a
non-zero value.

\begin{figure}
\includegraphics[width=\textwidth]{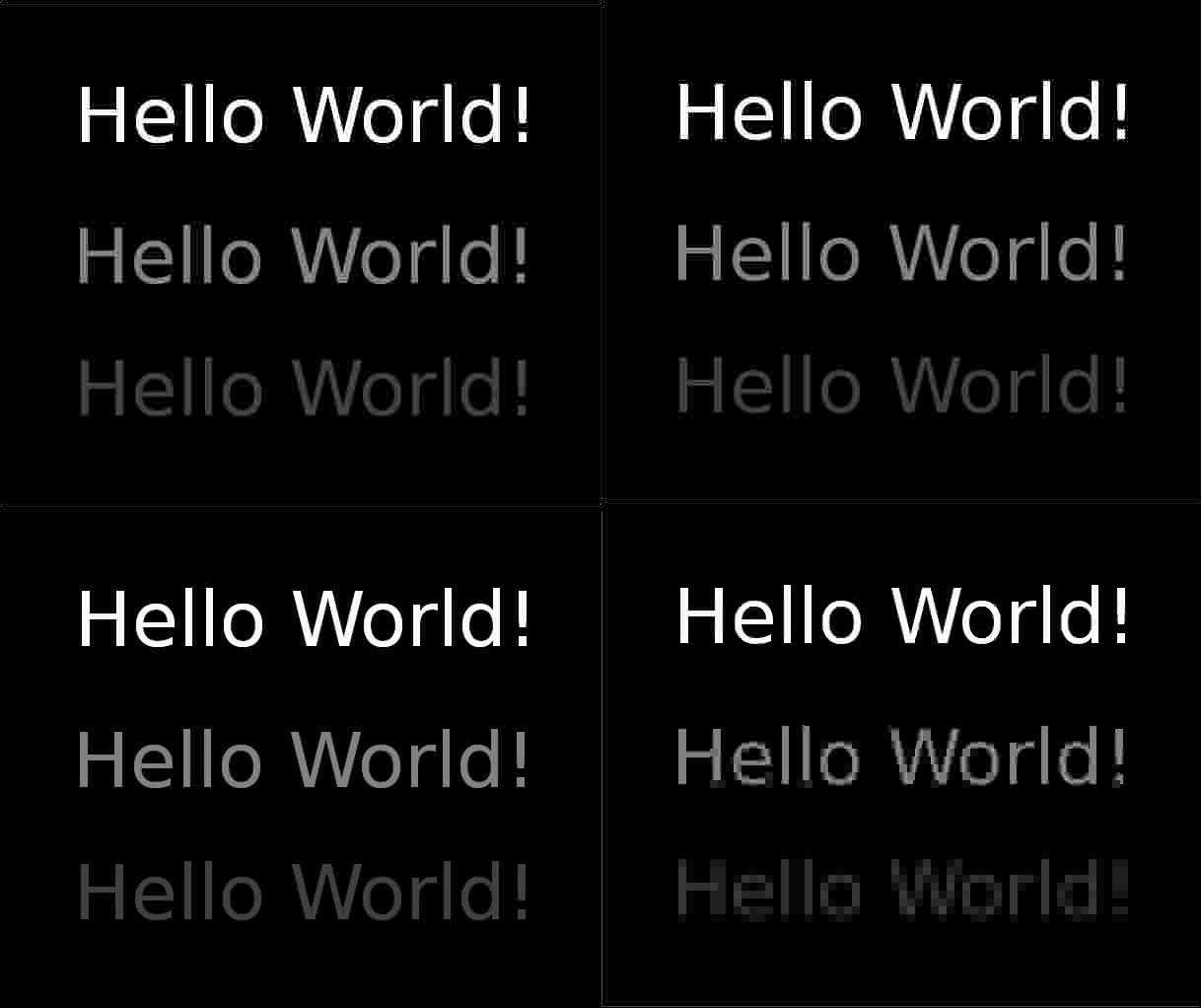}
\caption{The top part of the figure shows how \texttt{imagemagick convert --
subsample-factor 1x1x1} stores the blue channel of the image in
Figure~\ref{fig:blue-on-background}. One can observe that the blue channel,
which is displayed here as a grayscale image, is stored with similar accuracy in
the black area and in the yellow area. \\
The bottom part of the figure shows how Guetzli stores the blue channel of the
image in Figure~\ref{fig:blue-on-background}. Guetzli compromises the quality of
low blue values more when the blue perception is masked by yellow, but stores it
with higher accuracy when blue modulation is on the black background.}
\label{fig:blue-distortion}
\end{figure}

Lastly, we consider an encoding in YUV420 mode, where two out of three channels
are downsampled by $2\times{}2$.  Unfortunately, YUV420's handling of an area of the
image does not depend on the colors involved, and so it cannot capture effects
such as the one in Fig.~\ref{fig:blue-on-background} (see Fig.~\ref{fig:blue-distortion}
for the distortion that Guetzli applies to that image, which still
cannot be seen). In many cases encoding an image in YUV420 mode, with no
quantization, already causes a visible distortion. Thus, YUV420 is rarely useful
in the quality range Guetzli targets.

We have also tried to get space savings by decreasing (but not zeroing out)
absolute values of some coefficients. We hoped that by doing that we can decrease
the size of a coefficient at a lower cost to the image quality. However, we
could not find a way to beneficially combine it with zeroing out of the
coefficients.

We have also tried to modify the coefficients to compensate for distortions
caused by previously-mentioned optimizations. In order to find the compensating
modifications, we've computed the derivative of an approximation of Butteraugli.
Unfortunately, such modifications are usually smaller than quantization
intervals, so they cannot be applied.

We chose to forego some options due to undesired effects they would have. We
have chosen not to resample the image to a lower resolution. Using a lower
spatial resolution is often a practical approach~\cite{romano2016raisr}, but we
left it out from automated optimizations as we thought that it is somewhat
orthogonal to the optimizations we do and can be implemented as a higher-level
optimization. We have also decided to forego producing progressive JPEGs (we
always produce sequential JPEGs). Although progressive JPEGs are $2$-$5\%$ smaller,
they are 17-200\% slower to decode~\cite{progressive-performance}.

\subsection{Optimization procedure}

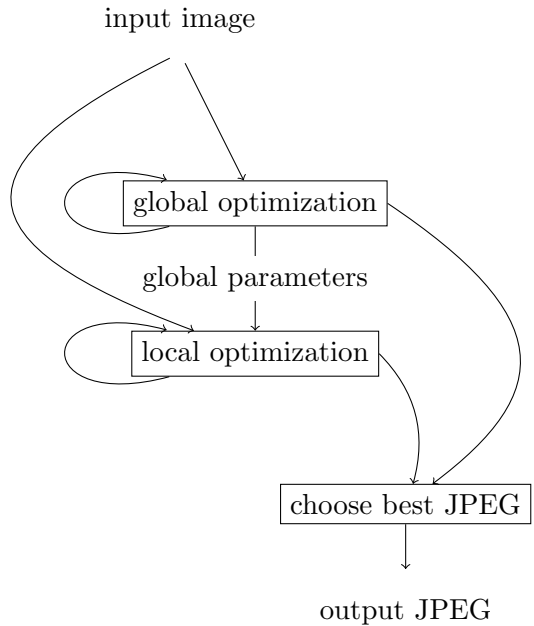
\begin{figure}
\begin{tikzpicture}
\tikzstyle{box} = [draw, rectangle]
\node[label=above:input image] (input) at (-1,1) {};
\node[box] (global) at (0,-1) {global optimization};
\node[box] (local) at (0,-3) {local optimization};
\node[box] (select) at (2,-5) {choose best JPEG};
\node[label=below:output JPEG] (output) at (2,-6) {};
	\draw (input) edge[->] (global);
	\draw (global) edge[->, loop left] (global);
	\draw[->] (input) .. controls (-4,-0.5) and (-4, -1.5) .. (local);
	\draw (global) edge[->] node [midway, fill=white] {global parameters} (local);
	\draw (local) edge[->, loop left] (local);
	\draw[->] (global.east) .. controls (4,-2.5) and (4,-3.5) .. (select);
	\draw (local.east) edge[bend left,->] (select);
	\draw (select) edge[->] (output);

\end{tikzpicture}
\caption{High-level overview of Guetzli operation. Guetzli first tunes global
parameters and only then tunes local parameters, holding global
parameters constant. During the whole course of tuning, candidate JPEGs
are generated and the best one of them is chosen as the final output.}
\label{fig:diagram}
\end{figure}

Guetzli uses an iterative optimization process. In order to make the problem
simpler, the optimizer is not guided by file size. Instead, it is driven by the
perceptual quality target alone. It aims to create a JPEG encoding with
perceptual distance below a given threshold, as close to the threshold as possible.
Each iteration produces a candidate output JPEG and, at the end, the best (not
necessarily the last) one of them is selected. As mentioned previously, there
are two adjustments we can make to the image: global ones (quantization table)
and local ones (replacing coefficients with zeros). We make them in order
(Fig.~\ref{fig:diagram}):  We first generate some number of proposals while
tuning the global adjustments only. At the same time we decide on the final set
of global adjustments. Then, while using the final set of global adjustments, we
generate more proposals while tuning the local adjustments.

\subsubsection{Global quantization table optimizations}

Changes to the global quantization table impact the distortion of the whole image,
usually in a different manner and with a different magnitude in different areas.
A global quantization table is an array of 192 values. It is infeasible to
perform anything that approximates an exhaustive search of that space. Instead,
we have selected a set of predefined quantization tables and we use tables from
that set only.

We try to find a quantization table in that set that will produce psychovisual
distance not larger than $\alpha < 1$ times the desired maximal distance, when
no other distortions are applied. The multiplication factor $\alpha$ was chosen
experimentally to be $0.97$, because this value yielded the smallest possible
final output images. This small amount of slack allows for local optimizations
to be done everywhere in the image.

\subsubsection{Replacing individual coefficients with zeros}

The JPEG format is extremely efficient in coding zero DCT coefficients, as it
has a joint-entropy RLE approach to coding zeros together with DCT value
prefixes. At the same time cost savings achievable by any other coefficient
modification are much smaller. In practice, the more zeros we have, the smaller
the resulting JPEG. Because of this, much of Guetzli's power depends on
choosing the correct coefficients to zero out.

The choice of coefficients to zero out in far away blocks is essentially
independent. However, the choice in nearby blocks has to be coordinated for
two distinct reasons. Most obviously, the choices in both blocks that share an
edge impact artifacts on that edge. In some cases it appears that introducing
distortion on both sides of the edge causes the distortion to be less visible
than if one of the blocks was unchanged. Secondly, the visual impact of many
small artifacts in the same vicinity is additive.

In order to take these dependencies into account, we adjust the zeroing-out
choice in all blocks simultaneously.  Similarly to the global optimization
phase, after each such adjustment we produce a candidate output image. We then
compute the psychovisual distance between the original and candidate image and
use it to decide on the next adjustment. Thus the feedback loop, which uses a
block-agnostic distance metric, provides the required coordination across nearby
blocks.

In order to simplify the adjustments, we first determine the relative importance
of coefficients in each block.  This is done using a Butteraugli-derived
heuristic. We then zero out some number of the least important coefficients in
each block, according to our importance estimation. Before producing each
subsequent candidate, we simply adjust the number of coefficients zeroed out so
that the psychovisual error is below the threshold, as close as possible to the
threshold, everywhere.

The zeroing out is by far the most powerful part of Guetzli. The reduction in size
gained by using global and local optimization over just using local optimization
(with some reasonable defaults for quantization tables) is only about $10\%$.

\section{Results}

We evaluate the Guetzli compressor and compare its performance at the same
psychovisual distortion measured by Butteraugli. Butteraugli is the metric
Guetzli optimizes for. Thus, this experiment tests Guetzli's optimization
abilities and, in itself, does not measure visual quality of the results. We
will separately publish results of a human rating study that does compare visual
quality, as perceived by humans.

Our image corpus (\cite{corpus}) has been created by taking photos with a Canon
EOS 600d camera, storing them using highest quality JPEG settings and
downsampling the resulting images by $4\times{}4$ using Lanczos resampling, as
implemented in GIMP. Some of the images had unsharp masking applied to them
before the $4\times{}4$ resampling.

We have compared Guetzli to libjpeg and mozjpeg. We ran libjpeg and mozjpeg at
quality 95, both with and without chroma downsampling. Additionally, for mozjpeg
we have tried three values of the tune parameter (hvs-psnr, ssim and ms-ssim).
The procedure we used to generate the results is detailed in
Algorithm~\ref{lst:comparison}. The script used to implement that procedure can
be found at \url{https://goo.gl/jON2lC}.

\begin{algorithm}
  \ForEach{other compressor, settings of other compressor}{
    \ForEach{image}{
      Compress the image with the other compressor and measure the Butteraugli
      distance. Let us call this JPEG file the \emph{other compressor's
      JPEG.}\\\vspace{0.1in}
      Compress the image with Guetzli targeting the Butteraugli distance measured
      in the previous step. Let us call this JPEG file the \emph{Guetzli JPEG.}
    }
    Compute total size of other compressor's JPEGs.\\
    Compute total size of Guetzli JPEGs.
  }
  \caption{Comparison procedure}
  \label{lst:comparison}
\end{algorithm}

\begin{table}
\small\begin{tabular}{lrrr}
\multirow{2}{*}{JPEG Encoder} & \multicolumn{2}{c}{Corpus size (bytes)} &
\multirow{2}{*}{Savings} \\
& Other encoder\textsuperscript{\textdagger} & Guetzli\textsuperscript{*} & \\

libjpeg -quality 95 &
5\,197\,681&
2\,952\,897&
-43.19\%\\
libjpeg -sample 1x1 -quality 95&
7\,049\,784&
4\,639\,276&
-34.19\%\\
mozjpeg -quality 95&
4\,195\,574&
2\,968\,525&
-29.25\%\\
mozjpeg -sample 1x1 -quality 95&
6\,502\,510&
4\,177\,354&
-35.76\%\\
mozjpeg -quality 95 -tune-ssim&
6\,498\,433&
3\,740\,070&
-42.45\%\\
mozjpeg -sample 1x1 -quality 95 -tune-ssim&
10\,133\,646&
6\,721\,080&
-33.68\%\\
mozjpeg -quality 95 -tune-ms-ssim&
4\,122\,235&
2\,775\,398&
-32.67\%\\
mozjpeg -sample 1x1 -quality 95 -tune-ms-ssim&
6\,481\,197&
3\,539\,223&
-45.39\%\\
mozjpeg -quality 95 -baseline&
4\,375\,890&
2\,968\,554&
-32.16\%
\end{tabular}
\caption{Comparison with other JPEG encoder at same Butteraugli distances. \\
\textsuperscript{\textdagger}---Size of corpus compressed using the other
encoder \\
\textsuperscript{*}---Size of corpus compressed with Guetzli at a quality that
matches the other encoder's Butteraugli distance
}
\label{tab:results}
\end{table}

The results are summarized in Table~\ref{tab:results}. When comparing Guetzli to
another compression algorithm at the same Butteraugli score, we can reach
savings in size between of 29-45\% savings in file size.

\section{Discussion}

The JPEG format does not support spatially adaptive quantization -- the
quantization arrays are constant across the whole image. However, one can
simulate this by creating more zeroes in areas with intended coarser
quantization. Using this poor substitute for adaptiveness, we can get partial
benefit out of exploiting visual masking phenomena (Fig.~\ref{fig:winnowing}). We
also use it to approximate nonuniform quantization in sRGB space. Format-level
support for spatially adaptive quantization and different colorspaces and/or
value-based adaptive quantization would make these optimizations much simpler
and more powerful.

It can be interesting to note that the effect in Fig.~\ref{fig:visual-masking}
is not captured in the JPEG format. The colorspace conversion from RGB to YUV is
linear, and cannot reduce accuracy of blue in a certain color mixtures.
Similarly, the change between YUV444 and YUV420 does not make a difference in
the representation accuracy of blue in different color environments -- the
spatial reduction happens similarly in black and yellow backgrounds. This
partially explains why YUV420 artifacts can be easily observable in images with
color details and dark backgrounds.

The results presented in this paper do not provide direct evidence of perceived
visual quality of results. We will separately publish results of a human rating
study designed to provide such evidence.

Some of our results are apples to oranges comparison as we compare progressive
JPEGs (mozjpeg) against sequential JPEGs (Guetzli). This puts Guetzli at a
disadvantage, but in a Butteraugli-based measurement we still get overall
savings for Guetzli ($-29.95\%$). It is questionable whether the savings at
transfer time are worth the slowdown at decoding time. We did not make an
attempt at progressive encoding with Guetzli, so we cannot be sure how much
smaller such images would be, but very likely we would get significant further
size savings from progressive encoding.

As with zopfli~\cite{zopfli}, our similar effort for the gzip/deflate/PNG format, Guetzli
is rather slow to encode. Getting a significant savings on static image content on
popular image heavy websites can be a possible actual use case. Although Guetzli
may be too slow for many practical uses, we hope that it can show direction for
future image format design.

We have shown that even despite the deficiencies of the JPEG format, we can
still greatly benefit from a complex psychovisual score such as Butteraugli,
and the approach we have chosen produces significantly smaller (29-45\%) file
sizes at a given psychovisual error score. The same approach can be applied to
a format that lacks these deficiencies (e.g. allows spatial adaptive
quantization, admits a richer description of quantization that can capture the
effect from Fig.~\ref{fig:visual-masking}) at a much smaller computational cost
and, likely, significantly larger compression ratio benefit.

{
\small
\bibliographystyle{unsrtnat}
\bibliography{references}{}
}

\end{document}